\documentclass[runningheads]{llncs}

\usepackage{eccv}

\usepackage{eccvabbrv}

\usepackage{graphicx}
\usepackage{booktabs}

\usepackage{multirow}
\usepackage{tabularx}
\newcolumntype{Y}{>{\centering\arraybackslash}X}
\newcolumntype{C}[1]{>{\centering}p{#1}}
\newcolumntype{Z}{>{\raggedleft\arraybackslash}X}

\usepackage[accsupp]{axessibility}  

\usepackage{hyperref}

\usepackage{orcidlink}

\usepackage{amsfonts}
\makeatletter
\newcommand*\tablesmallsize{%
  \@setfontsize\tablesmallsize{8}{9}%
}
\makeatother

\newcommand{\ds}{{SpaRe}}

\begin{document}

\title{AIM 2024 Sparse Neural Rendering Challenge:\\Methods and Results} 

\titlerunning{Sparse Neural Rendering Challenge}

\author{Michal Nazarczuk\thanks{Michal Nazarczuk \email{[michal.nazarczuk1@huawei.com]}, Sibi Catley-Chandar, Thomas Tanay, Richard Shaw, Eduardo Pérez-Pellitero (Huawei Noah’s Ark Laboratory), and Radu Timofte (University of Würzburg) are the Spare Neural Rendering Challenge organisers, while the other authors participated in the challenge. Appendix A contains the authors and their affiliations.} \and Sibi Catley-Chandar$^{\star}$ \and Thomas Tanay$^{\star}$ \and Richard Shaw$^{\star}$ \and Eduardo Pérez-Pellitero$^{\star}$ \and Radu Timofte$^{\star}$ \and Xing Yan \and Pan Wang \and Yali Guo \and Yongxin Wu \and Youcheng Cai \and Yanan Yang \and Junting Li \and Yanghong Zhou \and P. Y. Mok \and Zongqi He \and Zhe Xiao \and Kin-Chung Chan \and Hana Lebeta Goshu \and Cuixin Yang \and Rongkang Dong \and Jun Xiao \and Kin-Man Lam \and Jiayao Hao \and Qiong Gao \and Yanyan Zu \and Junpei Zhang \and Licheng Jiao \and Xu Liu \and Kuldeep Purohit
}

\institute{\,\vspace{-1em}}
\authorrunning{M.~Nazarczuk et al.}


\maketitle
\begin{abstract}
    This paper reviews the challenge on Sparse Neural Rendering that was part of the Advances in Image Manipulation (AIM) workshop, held in conjunction with ECCV 2024. This manuscript focuses on the competition set-up, the proposed methods and their respective results. The challenge aims at producing novel camera view synthesis of diverse scenes from sparse image observations. It is composed of two tracks, with differing levels of sparsity; 3 views in Track 1 (very sparse) and 9 views in Track 2 (sparse). Participants are asked to optimise objective fidelity to the ground-truth images as measured via the Peak Signal-to-Noise Ratio (PSNR) metric. For both tracks, we use the newly introduced \textbf{Spa}rse~\textbf{Re}ndering~(SpaRe) dataset \cite{Nazarczuk2024_dataset} and the popular DTU MVS dataset \cite{aanaes2016}.
    In this challenge, 5 teams submitted final results to Track 1 and 4 teams submitted final results to Track 2. The submitted models are varied and push the boundaries of the current state-of-the-art in sparse neural rendering. A detailed description of all models developed in the challenge is provided in this paper.

\end{abstract}
\section{Introduction}
The seminal work of Mildenhall \etal~\cite{Mildenhall2020} introduced Neural Radiance Fields (NeRF) and pioneered the use of implicit neural functions representing the 3D geometry and radiance of a scene, supervised with dense posed imagery via volumetric differentiable rendering.
This novel approach obtains impressive photorealistic results on the novel view synthesis task, especially when very dense view coverage of the scene is available.

In recent years, we have witnessed a bustling research community addressing a variety of open challenges and related applications, with major breakthroughs in \eg rendering speed and training time \cite{mueller2022, yu2021plenoctrees, lin2022enerf, Chen2022ECCV }, reconstruction accuracy \cite{kaizhang2020, barron2021, barron2023}, editing \cite{hyung2023local, bao2023sine, wang2022clip}, rasterisation paradigms \cite{kerbl2023}. Despite this remarkably fast progress, one of the key remaining challenges shared among a vast majority of the methods is the high sensitivity to the number of training views available, \ie reconstruction accuracy degrades quickly when only a handful of views are available \cite{niemeyer2022}.

Reconstructing a scene with a sparse set of input images is particularly challenging because it is at the core of the shape-radiance ambiguity, \ie models can easily explain the few image observations of the scene by fitting the wrong geometry. Prior art has made progress on sparse reconstruction with diverse approaches, \eg: generalisable methods aggregate prior knowledge by pretraining \cite{zhou2018stereo,mildenhall2019local, wang2021ibrnet, tanay2024}, depth regularisation and supervision \cite{niemeyer2022, wang2023, xu2024b}, appearance regularisation methods \cite{jain2021, wynn2023}. We refer the reader to \cite{Nazarczuk2024_dataset} for a more exhaustive taxonomy and related work review on sparse reconstruction methods.

The AIM 2024 Sparse Neural Rendering Challenge aims at stimulating
research for sparse-view neural rendering. Our proposed dataset \cite{Nazarczuk2024_dataset} and evaluation protocol are created to homogenise existing benchmarks and better understand the state-of-the-art landscape for different levels of sparsity in the input image set. This challenge is one of the AIM 2024 Workshop\!~\footnote{\url{https://www.cvlai.net/aim/2024/}} associated challenges on: sparse neural rendering, UHD blind photo quality assessment~\cite{aim2024uhdbpqa}, compressed depth map super-resolution and restoration~\cite{aim2024cdmsrr}, raw burst alignment~\cite{aim2024rawburst}, efficient video super-resolution for AV1 compressed content~\cite{aim2024evsr}, video super-resolution quality assessment~\cite{aim2024vsrqa}, compressed video quality assessment~\cite{aim2024cvqa} and video saliency prediction~\cite{aim2024vsp}.

\section{Challenge}

The AIM 2024 Sparse Neural Rendering Challenge addresses the task of novel view synthesis under sparse input constraints. The challenge aims to assess and advance state-of-the-art methods in sparse neural rendering. The focus of the challenge is on fair and up-to-date evaluation for sparse rendering. 

\subsection{Dataset}
For this challenge, we propose a new dataset that builds from the set-up of DTU, which is one of the most commonly used datasets for sparse reconstruction evaluation in the literature. In our dataset, we introduce new scenes for both training and testing of algorithms, and additionally reevaluate and refresh existing benchmarking protocols. 

Our new dataset, the \ds{} dataset~\cite{Nazarczuk2024_dataset} consists of $82$ training, $6$ validation, and $9$ test scenes. Each scene is composed of up to 9 input images, accompanied by input and target camera poses. Additionally, the training split of the dataset includes ground truth images for all camera poses enabling their use for model pre-training. 

The data used in the challenge is composed in part of the \ds{} dataset and in part of the existing DTU\,\cite{aanaes2016} scenes. We evaluate all images at full resolution ($1600\text{x}1200$) unlike previous works \cite{yu2021} which use resized images ($400\text{x}300$). Additionally, in contrast to prior works, we randomly select input and target camera views instead of using fixed poses throughout evaluation. Further details on the dataset and benchmark design are available in \cite{Nazarczuk2024_dataset}.

\subsection{Challenge Design and Tracks}

The challenge focuses on developing novel view synthesis solutions given the sparse input. To this end, we run the challenge in 2 tracks:
\begin{itemize}
    \item \textbf{Track 1:} $3$ input views per scene (\textit{very sparse}). This track provides very scarce input views with a limited amount of covisible regions of the object in the scene. This poses a significant challenge for the off-the-shelf neural reconstruction methods that often requires some form of regularisation to prevent over-fitting (\ie placing input views directly in front of the camera, failing to reconstruct underlying geometry).
    \item \textbf{Track 2:} $9$ input views per scene (\textit{sparse}). This track explores a less stringent sparse set-up, while still being an order of magnitude more sparse than common set-ups. The use of $9$ input views introduces more shared cues between views, yet still is very challenging for dense reconstruction approaches \cite{Mildenhall2020, barron2021, barron2022, barron2023}. This track essentially reproduces an evaluation set-up commonly used in prior art, firstly proposed in \cite{yu2021}). 
\end{itemize}

\subsection{Challenge Phases}
The challenge consists of two distinct phases, a development phase intended to allow participants to improve and validate their models, and a testing phase designed to evaluate the final submission. 
\subsubsection{Development Phase}
Participants are provided with the validation split of the data, including input view images and target poses, and the training split which includes full ground truth images for all camera poses. The participants are able to compute all fidelity metrics by submitting the predicted target views into the Codalab challenge server. The leaderboard is visible to all participants. In the development phase, the participants are provided with a baseline approach as a starting step and a sanity check for the submission system.
\subsubsection{Final Phase}
Participants are provided with the test split of the data, namely input view images and target poses. In this phase, some scenes are shared across Track 1 and and Track 2 while other scenes are unique to a single track only. This is to ensure the detectability of potential cross-contamination from the $9$ view track to the $3$ view track. Unlike in the development phase, the final phase results and leaderboard remain hidden from the participants. Additionally, all participants are asked to provide the factsheet documenting the solution and the code used to generate submitted predictions. Once the phase is over, the organisers run and validate the code to obtain the final results. 

\subsection{Evaluation}

The evaluation of the challenge is based on several image quality metrics. Firstly, we use the well-known standard peak signal-to-noise ratio (PSNR). We compute this metric both on the whole image (PSNR) and also only within the mask of the object in the scene (PSNR-M). From these two, we select PSNR-M as the primary metric to rank methods in the challenge as we put more emphasis on object reconstruction than background reconstruction.

Further, we provide additional image quality metrics. We calculate the Structural Similarity Index Measure (SSIM)~\cite{wang2004} within a tight bounding box around the object mask (SSIM-M). Similarly, we provide the Learned Perceptual Image Patch Similarity (LPIPS)~\cite{zhang2018} calculated in the bounding box (LPIPS-M).
\section{Teams and Methods}

\subsection{wang\_pan}

\begin{figure}
    \centering
    \includegraphics[width=0.8\linewidth]{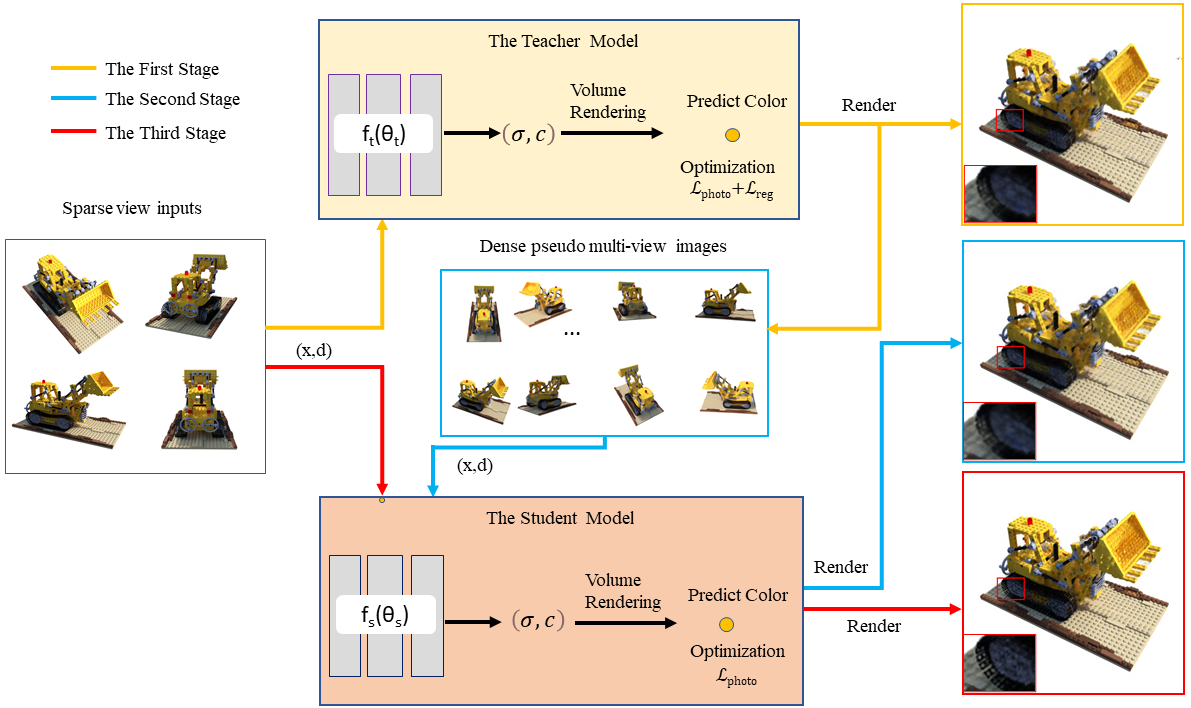}
    \caption{An overview of FrameNeRF~\cite{xing2024} proposed by team \textit{wang\_pan}.}
    \label{fig:wang_pan_overview}
\end{figure}

The team proposes FrameNeRF~\cite{xing2024}, an approach based on two models serving as teacher and student. The teacher model handles sparse input images and learns coarse scene geometry. The student model learns high-quality reconstruction from the provided input whilst being regularised through pseudo-groundtruth views produced by the teacher. An overview of their method is shown in Fig.~\ref{fig:wang_pan_overview}.

The method consists of three main steps. Firstly, the teacher model is trained on sparse input views to learn the coarse geometry of the scene. Here, FreeNeRF~\cite{yang2023} is used as the underlying model and is trained for 30K iterations at 1/4 resolution of $400 \times 300$px to reduce computational resources. The model is used to generate $49$ images of the object corresponding to dense multi-view coverage. Secondly, these images are used as pseudo-groundtruths to train the student model, refining the underlying structure of the scene. The student in this approach is based on Zip-NeRF~\cite{barron2023}, chosen for its high-quality reconstruction ability. The student is initially trained from dense teacher inputs only to regularise the geometry of the underlying 3D object. In this stage, the pseudo groundtruth images are upscaled back to full resolution and the student model is trained for 5K iterations. Finally, the student model is fine-tuned on the original sparse input images for 5K iterations, where the more accurate geometry from stage 2 enables more precise colour propagation to unobserved viewpoints and further optimises the scene geometry by removing existing teacher-induced artifacts such as floaters. Both tracks employ the same solution.

\subsection{MikeLee}

The team proposes a method adapted from Self-Conditioned NeRF (SCNeRF)~\cite{li2024_scnerf}, leveraging information from features extracted from pretrained networks to guide the training of radiance fields in the sparse-view setting. 
The overall framework, shown in Fig.~\ref{fig:MikeLee_fig1}, introduces two modifications:

\begin{figure}
    \centering
    \includegraphics[width=0.8\linewidth]{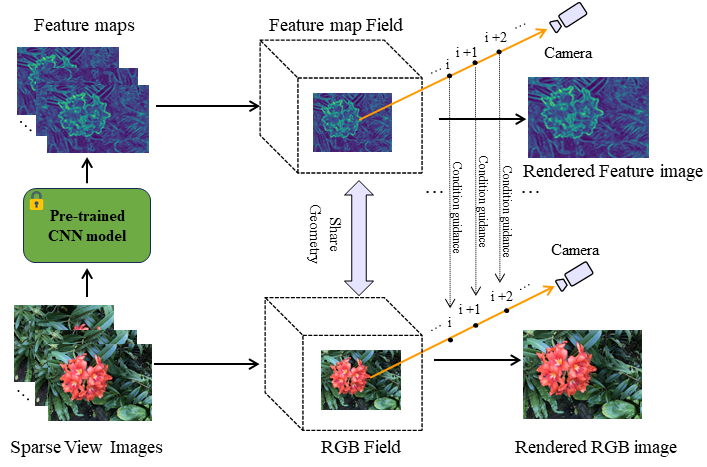}
    \caption{An overview of the method proposed by \textit{MikeLee}. The framework learns and combines information from two neural fields: one branch learns an RGB field, while the other learns a feature field, sharing geometry information. The colour prediction branch is conditioned on the prior learned from the feature branch. The network is trained to predict local features and colour at the pixel level in the sparse training views.
    }
    \label{fig:MikeLee_fig1}
\end{figure}

\begin{figure}
    \centering
    \includegraphics[width=0.8\linewidth]{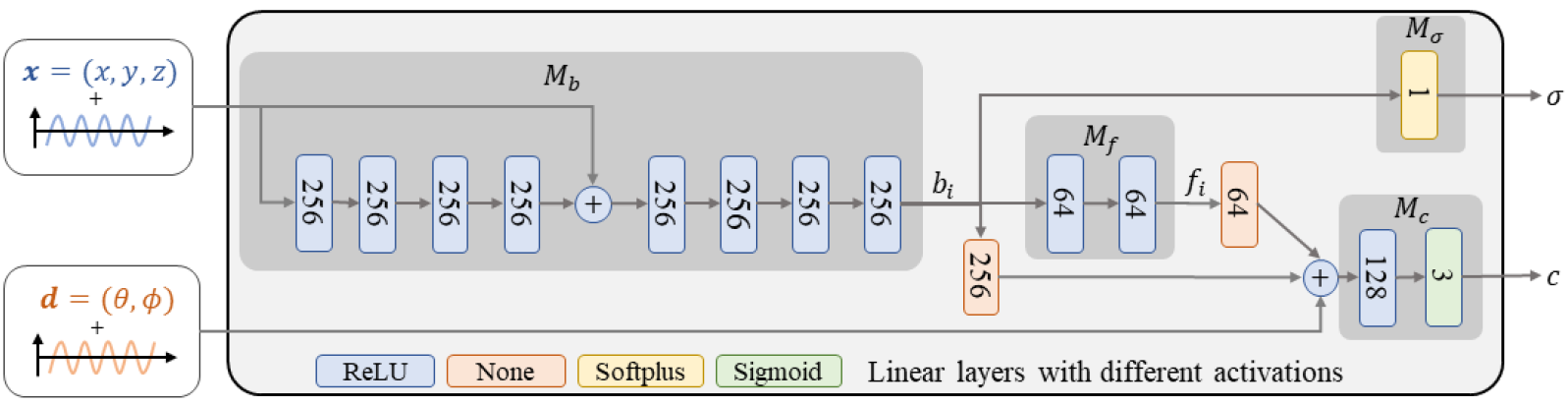}
    \caption{In the method proposed by \textit{MikeLee}, the network predicting the colour $c$ of a point is explicitly conditioned on the local features of the point. Feature supervision supervises $f_i$ based on prior knowledge from a pretrained network; Feature condition concatenates the learned prior $f_i$ as additional input to $M_c$ for colour prediction.}
    \label{fig:MikeLee_fig2}
\end{figure}

\textbf{Modification 1:} Local feature descriptors extracted from a pretrained network (VGG trained on ImageNet) are used to constrain the reconstruction process. For a 3D point on the surface of an object, its colour may have some variance when observed from different view directions, and in the sparse setting its colour loss is easy to overfit. However, the abstract description of the point should be similar from different views. DietNerf~\cite{jain2021} explored a similar idea that ``a [...] is a [...] from any perspective''. However, unlike DietNerf, which constrains the learning process in unobserved views with loss at the image level, here the learning process is supervised in the training views with loss at the pixel level, as shown in Fig.~\ref{fig:MikeLee_fig1}. Specifically, feature maps $F^{gt}$ are first extracted from the sparse input images using the pre-trained VGG model. Then, for a 3D point $p_i = (x, y, z)$, a network $M_b$ predicts a bottleneck feature $b_i$, as shown in Fig.~\ref{fig:MikeLee_fig2}, which is used by network branch $M_{\sigma}$ to predict density $\sigma_i$, independent of the view direction $d$:

\begin{align}
    b_i &= M_{b} (p_i) \\
    \sigma_i &= M_{\sigma} (b_i)
\end{align}

An additional MLP $M_f$ is used to predict prior feature $f_i$ for the 3D point based on the shared bottleneck feature $b_i$:

\begin{align}
    f_i &= M_f (b_i) .
\end{align}

As feature $b_i$ is input to both the density MLP $M_{\sigma}$ and feature MLP $M_f$, information is shared between the two branches. The feature $F$ at a corresponding pixel is obtained by volume rendering, and the distance between the rendered feature maps $F (r)$ and the extracted features $F^{gt} (r)$ for each ray $r$ is minimised with $L_2$ loss:

\begin{equation}
    L_f = || F (r) - F^{gt} (r) ||_2 .
    \label{eq:MikeLee_loss1}
\end{equation}

\textbf{Modification 2:} The relation of feature descriptors and colour predictions is explicitly constrained. For a 3D point on the surface an object, it's colour should have a high probability of being of similar colour to rest of the object. In a similar approach to \textit{Distilled feature fields} (DFFs)~\cite{Kobayashi2022}, which showed that by learning colours and features of 3D points simultaneously NeRF can decompose the scene into different semantic parts (objects), here the method conditions the NeRF as in Fig.~\ref{fig:MikeLee_fig2}, so that the feature learning can also benefit the colour learning. Specifically, the bottleneck feature $b_i$, prior feature $f_i$ and view direction $d$ are concatenated and fed into MLP $M_c$ to predict the final pixel colour $c_i$:

\begin{align}
    c_i = M_c (b_i, f_i, d)
\end{align}

The overall loss function is then the sum of feature and colour losses:

\begin{equation}
    L = L_c + \lambda L_f
    \label{eq:MikeLee_loss2}
\end{equation}

\noindent where $\lambda$ is the balancing weight for the feature loss.

The pre-trained VGG network extracts features and the Relu1-1 layer is used for feature supervision (Eq.~\ref{eq:MikeLee_loss1}). This layer's output provides a description of a pixel's local neighbourhood, and although it does not contain high-level abstract information about the object, it contains some prior information. This layer is the same size as the input image and thus will not introduce interpolation artifacts compared to deeper layers. The two modifications are applied to FreeNeRF~\cite{yang2023} and the network is trained at low resolution for 44K iterations, with frequency regulation ending at 40K iterations. All other parameters are kept to default. The batch size is set to 1024 due to memory limitations. The team only took part in Track 1 of the challenge. 

\subsection{zongqihe}

\begin{figure}
    \centering
    \includegraphics[width=0.8\linewidth]{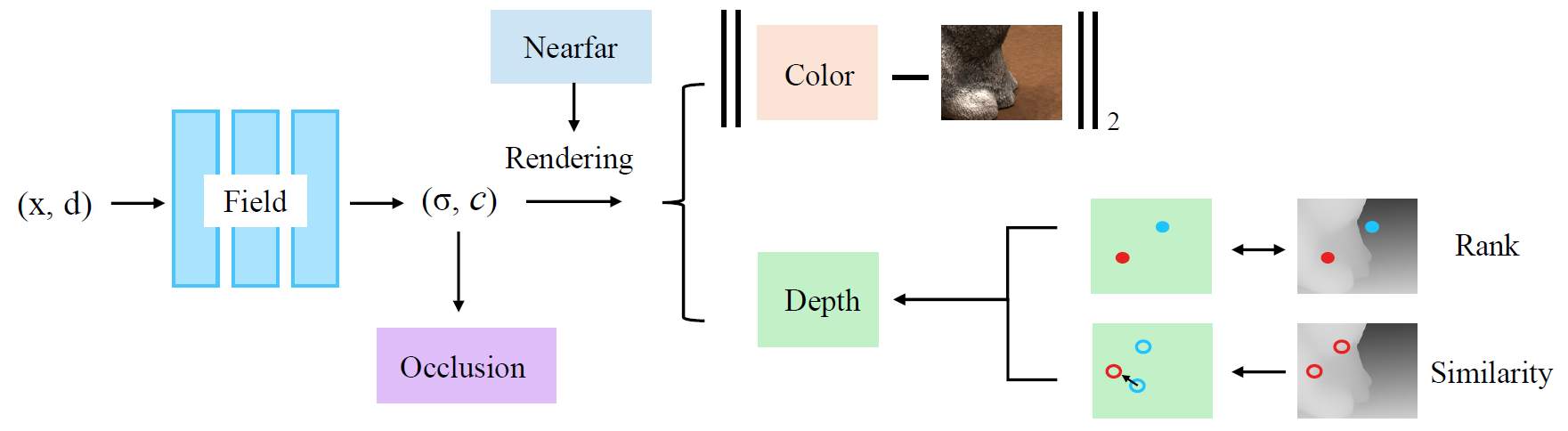}
    \caption{An overview of ESNeRF proposed by \textit{zongqihe}. Colour- and depth-based losses are applied, in addition to ``occlusion'' regularisation and near-far field optimisation.}
    \label{fig:zongqihe_overview}
\end{figure}

The team proposes ESNeRF (Extremely Sparse Neural Radiance Fields), incorporating pixel-~\cite{Mildenhall2020} and depth-based losses~\cite{wang2023}, leveraging depth information generated through a pretrained model, \ie DPT~\cite{Ranftl2021VisionTF}, for supervision. Fig.~\ref{fig:zongqihe_overview} presents the overall framework. Due to the ill-posed nature of novel view synthesis in the sparse-view setting and to address issues such as overfitting, a hybrid loss function is proposed:

\begin{equation}
    L_{total} = L_{NeRF} + w_1 L_{TV} + w_2 R_{rank} + w_3 R_{cont}.
\end{equation}

FreeNeRF~\cite{yang2023} is used as the backbone model, with colour reconstruction loss $L_{NeRF}$ defined for a set of rays $\mathcal{R}$ as:

\begin{equation}
    L_{NeRF} = \sum_{r \in \mathcal{R}} \left\| \hat{C}_c(r) - C(r) \right\|^2 + \left\| \hat{C}_f(r) - C(r) \right\|^2 ,
\end{equation}

\noindent where and $C(r)$, $\hat{C}_c(r)$, $\hat{C}_f(r)$ are the groundtruth pixel colour, coarse and fine rendered colours for ray $r$ respectively. As the depth maps generated by DPT may be inaccurate and lack sufficient detail, using them solely to supervise the NeRF deteriorates the render quality. Therefore, three additional regularisation losses are introduced into the training process:

\textbf{Total Variation Loss:} 
To avoid abrupt changes between neighbouring values of rendered depth, a depth variance loss $L_{TV}$ computing depth variance relative to neighbouring pixels, promoting spatial consistency and smoothness, is defined:

\begin{equation}
    L_{TV} = \sum_{i,j} \left( |d_{i,j} - d_{i+1,j}|^2 + |d_{i,j} - d_{i,j+1}|^2 \right) ,
    \label{eq:zongqihe_loss1}
\end{equation}

\noindent where $d_{i,j}$ is the depth at pixel $(i,j)$, $d_{i+1,j}$ and $d_{i,j+1}$ are depths at the pixels directly to the right and below respectively.

\textbf{Depth-Guided Ranking Regularisation:} 
By comparing two random points from the pretrained model's depth map $d$ with the depth rendering $\hat{d}$, the model is constrained to maintain surface geometry consistency. 
Let $\mathcal{P}$ be a set of local patches extracted from the input image $I$, the depth-guided ranking regularisation is defined as follows:

\begin{align}
    R_{rank} =  \sum_{d_i \leq d_j} \max \left( \hat{d}_i - \hat{d}_j + k, 0 \right) ,
\end{align}

\noindent where $\hat{d} \in \mathcal{P}$ represents the local depth map, estimated by volume rendering, $\hat{d}_i$ and $\hat{d}_j$ are the $i$-th and $j$-th patches of predicted depths, respectively. The regularisation term penalises incorrect depth ordering of the predictions. Specifically, when two randomly sampled points from $d$ satisfy $d_i \leq d_j$, but the corresponding rendered depth violates the ordering consistency, \ie $\hat{d}_i > \hat{d}_j$, the penalty term guides the model to correct the depth ordering. The constant $k$ provides some tolerance to avoid penalising small depth ranking errors.

\textbf{Depth-Guided Continuity Regularisation:} Depth ranking helps the model learn a consistent depth representation, but alone cannot capture the geometric details of the scene. An additional depth-guided continuity regularisation term penalising large depth differences between neighbouring pixels is proposed:

\begin{align}
    R_{cont} =  \sum_i \sum_{d_j \in \mathrm{KNN}(d_i)} \max \left( | \hat{d}_i - \hat{d}_j | - k', 0 \right) ,
\end{align}

\noindent where for each pixel $i$, $K$ nearest neighbours $\mathrm{KNN} ( \cdot )$ are identified from the input depth map. The penalty term ensures that the difference between the predicted depth values $\hat{d}_i$ and $\hat{d}_j$ does not exceed a predefined threshold $k'$. 

In addition to the aforementioned losses, ``occlusion'' regularisation~\cite{yang2023} and near-far field optimisation are introduced during the rendering of rays to improve the accuracy of depth details. The weight of occlusion loss is set to 0.1. The weight $w_1$ of the total variation loss undergoes linear annealing, where at the maximum training step $w_1 = 1$, while $w_2$ and $w_3 = 0.2$. The model is trained for 10K iterations for all scenes.

\subsection{Thirteen}

\begin{figure}
    \centering
    \includegraphics[trim={0 3.5cm 2cm 0},clip,width=0.8\linewidth]{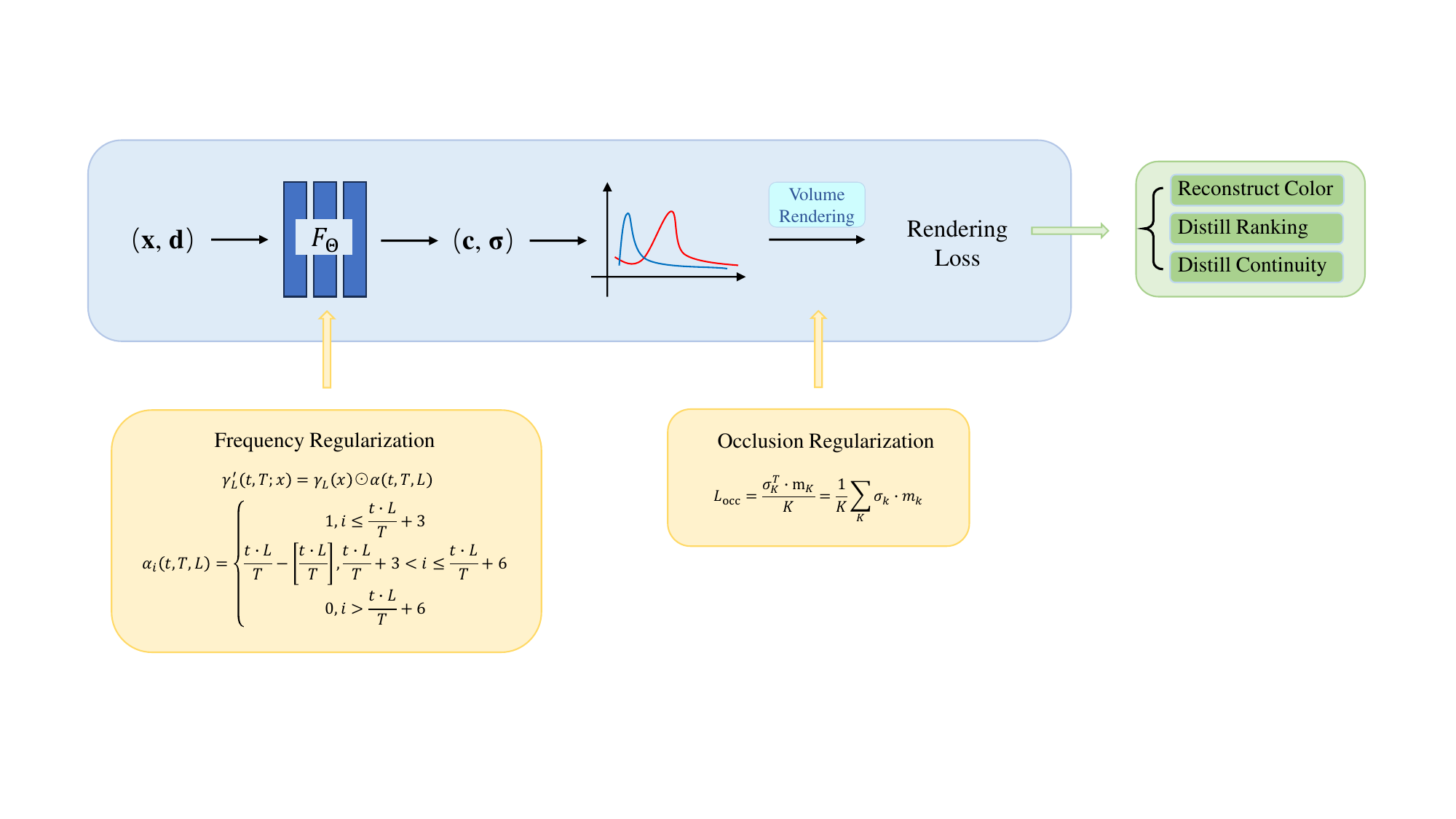}
    \caption{An overview of the method proposed by \textit{Thirteen}.}
    \label{fig:Thirteen_overview}
\end{figure}

The method provided by \textit{Thirteen} is divided into three models: baseline (FreeNeRF~\cite{yang2023}), SparseNeRF~\cite{wang2023}, and \textit{model fusion}. An overview is shown in Fig.~\ref{fig:Thirteen_overview}. 

Firstly, the team uses FreeNeRF which proposes two regularisation terms: one regularises the frequency range of NeRF’s inputs, while the other penalises near-camera density fields, thus improving few-shot neural rendering with no additional computational cost. The team use the frequency regularisation of FreeNeRF, while prior information of white and black backgrounds is used for occlusion regularisation on the DTU dataset. They train this model for 20K iterations. Secondly, SparseNeRF performs distilling depth ranking for fewshot novel view synthesis. The team integrate SparseNeRF into FreeNeRF and use the fused code to train the model on the DTU dataset with the same parameters. Finally, \textit{model fusion} fuses the results from the two prior phases. Two fusion methods are used: i) pixel-weighted fusion of the results generated by the different models, and ii) by evaluating and fusing the final results through SSIM and PSNR. They train their model on a single NVIDIA GeForce RTX 4090 GPU.

\subsection{IPCV}

\begin{figure}
    \centering
    \includegraphics[trim={0 8cm 9cm 7cm},clip,width=0.8\linewidth]{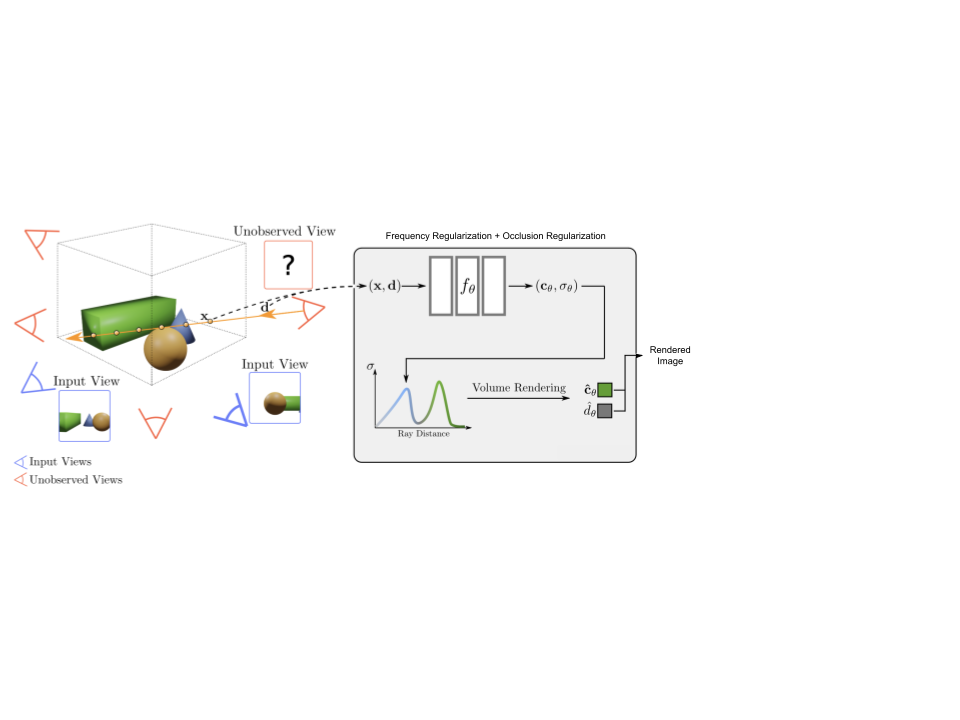}
    \caption{An overview of the method proposed by \textit{IPCV}.}
    \label{fig:kuldeeppurohit3_overview}
\end{figure}

This team's method is based on Freenerf~\cite{yang2023}. An overview is shown in Fig.~\ref{fig:kuldeeppurohit3_overview}.

\textbf{Frequency Regularisation:} The most common failure mode of few-shot neural rendering is overfitting the 2D images with small loss while not explaining 3D geometry in a multi-view consistent way. This is exacerbated by high-frequency inputs, therefore the team employs frequency regularisation to reduce overfitting caused by high-frequency inputs. Given a positional encoding $\gamma_L (x)$ of length $L + 3$, a linearly increasing frequency mask $\boldsymbol{\alpha}$ is used to regulate the visible frequency spectrum based on the training time steps, as follows:

\begin{equation}
  \begin{aligned}
    \gamma'_L &= \gamma_L (x) \odot \boldsymbol{\alpha} (t, T, L)\\
      \mathrm{with}\ \boldsymbol{\alpha}_i (t, T, L) & = 
      \begin{cases}
      1 & \text{if}\ i \leq \frac{t \cdot L}{T} + 3\\
      \frac{t \cdot L}{T} & \text{if}\ \frac{t \cdot L}{T} + 3 < i \leq \frac{t \cdot L}{T} + 6\\
      0 & \text{if}\ i > \frac{t \cdot L}{T} + 6
    \end{cases}     
  \end{aligned}
\end{equation}

\noindent where $\boldsymbol{\alpha}_i(t, T, L)$ denotes the $i$-th bit value of $\boldsymbol{\alpha}(t, T, L)$; $t$ and $T$ are the current and final iteration of frequency regularisation, respectively. Starting with raw inputs without positional encoding, the visible frequency linearly increases by 3-bit each time as training progresses. The frequency regularisation circumvents the unstable and susceptible high-frequency signals at the beginning of training and gradually provides NeRF high-frequency information to avoid over-smoothness.

\textbf{Occlusion Regularisation:} Due to the limited number of training views and ill-posed nature of the problem, certain characteristic artifacts may still exist in novel views. The presence of floaters and walls in novel views is caused by the imperfect training views, and thus can be addressed directly at training time without the need for novel-pose sampling. To this end, a simple yet effective ``occlusion'' regularisation is used to penalise the dense fields near the camera:

\begin{equation}
    \mathcal{L}_{occ} = \frac{\boldsymbol{\sigma}^T_K \cdot m_K}{K} = \frac{1}{K} \sum_K \sigma_k \cdot m_k,
\end{equation}

\noindent where $m_k$ is a binary mask vector that determines whether a point will be penalised, and $\boldsymbol{\sigma}_K$ denotes the density values
of $K$ points sampled along the ray in the order of proximity to the origin (near to far). To reduce solid floaters near the camera, the values of $m_k$ up to index $M$, termed as the regularisation range, are set to 1 and the rest to 0.

The team follow the experimental settings of FreeNeRF, using the same steps for both tracks. Training is done at 1/4 image resolution for 5K iterations per scene. Afterwards, the generated target views are bilinearly upsampled to full resolution.

\section{Results}

Out of the $50$ participants registered to Track 1, $6$ entered the final phase and submitted results to the server. Of those, $5$ submissions complied with the factsheet and code submission requirement. In Track 2, $37$ participants registered, and $4$ proceeded to submit results, factsheet, and code in the final phase. We report the final phase results in Table\,\ref{tab:track1test} and Table\,\ref{tab:track2test} respectively. 

\subsection{Main Ideas}

Most of the current works in the field of sparse novel view synthesis can be classified into two groups: methods that optimise underlying representation for each scene separately, or methods that propose a generalisable solution. In this challenge, all participants chose to propose algorithms belonging to the former group, \ie per-scene optimisation. 

All participants build their solution on top of the algorithm proposed by FreeNeRF\,\cite{yang2023}. Two of the teams that submitted the final solution and factsheet focused on regularisation techniques in order to deal with the underconstrained problem of having sparse input views. Those teams proposed the use of frequency and occlusion regularisation. Further, two more teams decided to leverage priors generated by a pretrained model for supervision. One suggested including depth-based losses (depth ranking and similarity) into optimisation based on pseudo-ground-truth generated by a depth estimation network. The other proposes the use of prior in the form of a feature map extracted with a pretrained network, \ie the semantic features of the same object should be similar from every viewing direction. Finally, one team proposes the use of a teacher-student approach, where the former is a model conditioned for sparse views and able to recover the underlying geometry, and the latter is a model characterised by a higher quality reconstruction. The teacher reconstructs the geometry and is used to generate dense pseudo-views which can be used to train the student.

\begin{table}
\tablesmallsize
\setlength{\tabcolsep}{2pt}
\begin{center}
\caption{Results of Track 1 - Test Phase. \textsuperscript{\textdagger}\textit{Incomplete submission due to lack of factsheet description, thus not ranked.}}
\label{tab:track1test}
\begin{tabularx}{\linewidth}{Xrrrrrrrrrrrr}
\toprule
\multirow{2}{*}{Place} & \multicolumn{3}{c}{PSNR-M} & \multicolumn{3}{c}{PSNR} & \multicolumn{3}{c}{SSIM-M} & \multicolumn{3}{c}{LPIPS-M}\\
\cmidrule(lr){2-4}\cmidrule(lr){5-7}\cmidrule(lr){8-10}\cmidrule(lr){11-13}
 & Avg & DTU & Syn
& Avg & DTU & Syn
& Avg & DTU & Syn
& Avg & DTU & Syn\\
\midrule
wang\_pan & $18.67$ & $18.50$ & $18.83$ & $17.98$ & $16.73$ & $19.23$ & $0.665$ & $0.591$ & $0.740$ & $0.395$ & $0.420$ & $0.369$ \\
MikeLee & $18.30$ & $18.16$ & $18.43$ & $18.18$ & $17.00$ & $19.36$ & $0.654$ & $0.584$ & $0.725$ & $0.515$ & $0.584$ & $0.447$ \\
zongqihe & $18.11$ & $18.39$ & $17.83$ & $16.82$ & $16.19$ & $17.44$ & $0.625$ & $0.545$ & $0.705$ & $0.592$ & $0.659$ & $0.526$ \\
Thirteen & $16.64$ & $14.96$ & $18.31$ & $17.13$ & $14.88$ & $19.38$ & $0.603$ & $0.490$ & $0.716$ & $0.585$ & $0.691$ & $0.479$ \\
IPCV & $15.58$ & $14.63$ & $16.54$ & $15.84$ & $13.94$ & $17.73$ & $0.559$ & $0.452$ & $0.667$ & $0.635$ & $0.709$ & $0.560$ \\
\midrule
Baseline & $15.28$ & $14.40$ & $16.17$ & $15.60$ & $13.68$ & $17.52$ & $0.556$ & $0.452$ & $0.660$ & $0.641$ & $0.718$ & $0.563$ \\
ZacharyXIAO\textsuperscript{\textdagger} & $18.04$ & $18.32$ & $17.76$ & $16.72$ & $15.97$ & $17.47$ & $0.627$ & $0.548$ & $0.707$ & $0.591$ & $0.658$ & $0.523$ \\
\bottomrule
\end{tabularx}
\end{center}
\end{table}

\begin{table}
\tablesmallsize
\setlength{\tabcolsep}{2pt}
\begin{center}
\caption{Results of Track 1 - Development Phase. \textsuperscript{\textdagger}-results were not verified due to lack of factsheet submission.}
\label{tab:track1dev}
\begin{tabularx}{\linewidth}{Xrrrrrrrrrrrr}
\toprule
\multirow{2}{*}{Place} & \multicolumn{3}{c}{PSNR-M} & \multicolumn{3}{c}{PSNR} & \multicolumn{3}{c}{SSIM-M} & \multicolumn{3}{c}{LPIPS-M}\\
\cmidrule(lr){2-4}\cmidrule(lr){5-7}\cmidrule(lr){8-10}\cmidrule(lr){11-13}
 & Avg & DTU & Syn
& Avg & DTU & Syn
& Avg & DTU & Syn
& Avg & DTU & Syn\\
\midrule
MikeLee & $19.13$ & $19.62$ & $18.63$ & $18.37$ & $16.26$ & $20.49$ & $0.612$ & $0.595$ & $0.629$ & $0.590$ & $0.625$ & $0.554$ \\
wang\_pan & $16.62$ & $17.36$ & $15.88$ & $17.03$ & $15.61$ & $18.45$ & $0.536$ & $0.522$ & $0.550$ & $0.522$ & $0.490$ & $0.554$ \\
zongqihe & $16.44$ & $17.02$ & $15.86$ & $16.65$ & $15.06$ & $18.24$ & $0.543$ & $0.525$ & $0.561$ & $0.659$ & $0.675$ & $0.643$ \\
IPCV & $15.40$ & $16.13$ & $14.67$ & $16.17$ & $14.51$ & $17.83$ & $0.524$ & $0.495$ & $0.552$ & $0.677$ & $0.697$ & $0.658$ \\
\midrule
Baseline & $16.73$ & $16.72$ & $16.73$ & $16.96$ & $15.41$ & $18.50$ & $0.538$ & $0.509$ & $0.568$ & $0.661$ & $0.681$ & $0.642$ \\
sunshine\_yyz\textsuperscript{\textdagger} & $16.67$ & $16.77$ & $16.57$ & $17.07$ & $15.24$ & $18.89$ & $0.544$ & $0.515$ & $0.573$ & $0.656$ & $0.673$ & $0.640$ \\
\bottomrule
\end{tabularx}
\end{center}
\end{table}

\begin{table}
\tablesmallsize
\setlength{\tabcolsep}{2pt}
\begin{center}
\caption{Results of Track 2 - Test Phase.}
\label{tab:track2test}
\begin{tabularx}{\linewidth}{Xrrrrrrrrrrrr}
\toprule
\multirow{2}{*}{Place} & \multicolumn{3}{c}{PSNR-M} & \multicolumn{3}{c}{PSNR} & \multicolumn{3}{c}{SSIM-M} & \multicolumn{3}{c}{LPIPS-M}\\
\cmidrule(lr){2-4}\cmidrule(lr){5-7}\cmidrule(lr){8-10}\cmidrule(lr){11-13}
 & Avg & DTU & Syn
& Avg & DTU & Syn
& Avg & DTU & Syn
& Avg & DTU & Syn\\
\midrule
wang\_pan & $24.51$ & $24.56$ & $24.46$ & $23.87$ & $23.79$ & $23.94$ & $0.784$ & $0.759$ & $0.808$ & $0.262$ & $0.267$ & $0.257$ \\
Thirteen & $21.59$ & $20.14$ & $23.04$ & $21.45$ & $19.73$ & $23.16$ & $0.649$ & $0.549$ & $0.749$ & $0.516$ & $0.628$ & $0.403$ \\
zongqihe & $21.09$ & $21.27$ & $20.92$ & $20.56$ & $20.35$ & $20.77$ & $0.641$ & $0.596$ & $0.687$ & $0.567$ & $0.610$ & $0.524$ \\
IPCV & $20.41$ & $20.03$ & $20.78$ & $20.42$ & $19.42$ & $21.43$ & $0.587$ & $0.526$ & $0.647$ & $0.571$ & $0.628$ & $0.514$ \\
\midrule
Baseline & $20.43$ & $19.99$ & $20.87$ & $20.61$ & $19.72$ & $21.49$ & $0.585$ & $0.522$ & $0.648$ & $0.569$ & $0.625$ & $0.512$ \\
\bottomrule
\end{tabularx}
\end{center}
\end{table}

\begin{table}
\tablesmallsize
\setlength{\tabcolsep}{2pt}
\begin{center}
\caption{Results of Track 2 - Development Phase. \textsuperscript{\textdagger}-results were not verified due to lack of factsheet submission.}
\label{tab:track2dev}
\begin{tabularx}{\linewidth}{Xrrrrrrrrrrrr}
\toprule
\multirow{2}{*}{Place} & \multicolumn{3}{c}{PSNR-M} & \multicolumn{3}{c}{PSNR} & \multicolumn{3}{c}{SSIM-M} & \multicolumn{3}{c}{LPIPS-M}\\
\cmidrule(lr){2-4}\cmidrule(lr){5-7}\cmidrule(lr){8-10}\cmidrule(lr){11-13}
 & Avg & DTU & Syn
& Avg & DTU & Syn
& Avg & DTU & Syn
& Avg & DTU & Syn\\
\midrule
wang\_pan & $22.42$ & $24.01$ & $20.83$ & $23.30$ & $22.22$ & $24.37$ & $0.655$ & $0.670$ & $0.639$ & $0.368$ & $0.324$ & $0.413$ \\
IPCV & $20.66$ & $22.12$ & $19.20$ & $21.76$ & $19.94$ & $23.59$ & $0.587$ & $0.563$ & $0.611$ & $0.594$ & $0.599$ & $0.590$ \\
\midrule
Baseline & $21.02$ & $22.29$ & $19.76$ & $22.20$ & $20.83$ & $23.57$ & $0.587$ & $0.562$ & $0.613$ & $0.591$ & $0.595$ & $0.587$ \\
sunshine\_yyz\textsuperscript{\textdagger} & $20.95$ & $22.59$ & $19.31$ & $22.30$ & $21.19$ & $23.41$ & $0.595$ & $0.579$ & $0.610$ & $0.590$ & $0.590$ & $0.589$ \\
\bottomrule
\end{tabularx}
\end{center}
\end{table}

\subsection{Top Results}
The quantitative results of the challenge for Track 1 Test and Development phase, and Track 2 Test and Development phase are presented in Tables \ref{tab:track1test}, \ref{tab:track1dev}, \ref{tab:track2test}, \ref{tab:track2dev} respectively. 
The visualisation of selected scenes and test views from the Test phase of both tracks can be seen in Figure~\ref{fig:results_syn_test} for the synthetic SpaRe dataset, and in Figure~\ref{fig:results_dtu_test} for the DTU dataset.

\subsubsection{Track 1}

The final classification of Track 1 (Test phase - Table \ref{tab:track1test}) reveals a very close competition between the top-scoring solutions. We observe the winner \textit{wang\_pan} to have performed the best in all object-oriented metrics. Notably, the team achieved the best score in a decisive metric - masked PSNR, with $0.37dB$ improvement over the runner-up, and the best score in perceptual similarity (LPIPS-M) with a large margin over the second-best score. It is worth noting that \textit{MikeLee} achieved the best PSNR calculated over the whole image, suggesting that their model is more suitable than others with respect to background reconstruction. 

\begin{figure}
    \centering
    \includegraphics[width=0.9\linewidth]{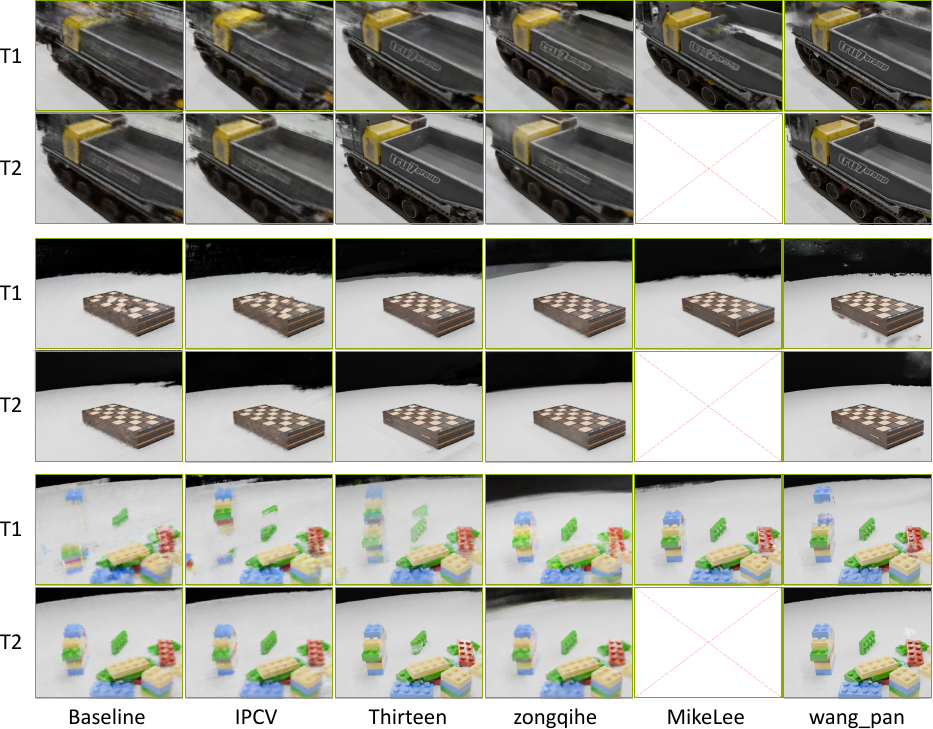}
    \caption{Test set results on the synthetic SpaRe dataset for Track 1 (T1) and Track 2 (T2). Ground truth images are omitted to preserve benchmark integrity.}
    \label{fig:results_syn_test}
\end{figure}

\begin{figure}
    \centering
    \includegraphics[width=1.0\linewidth]{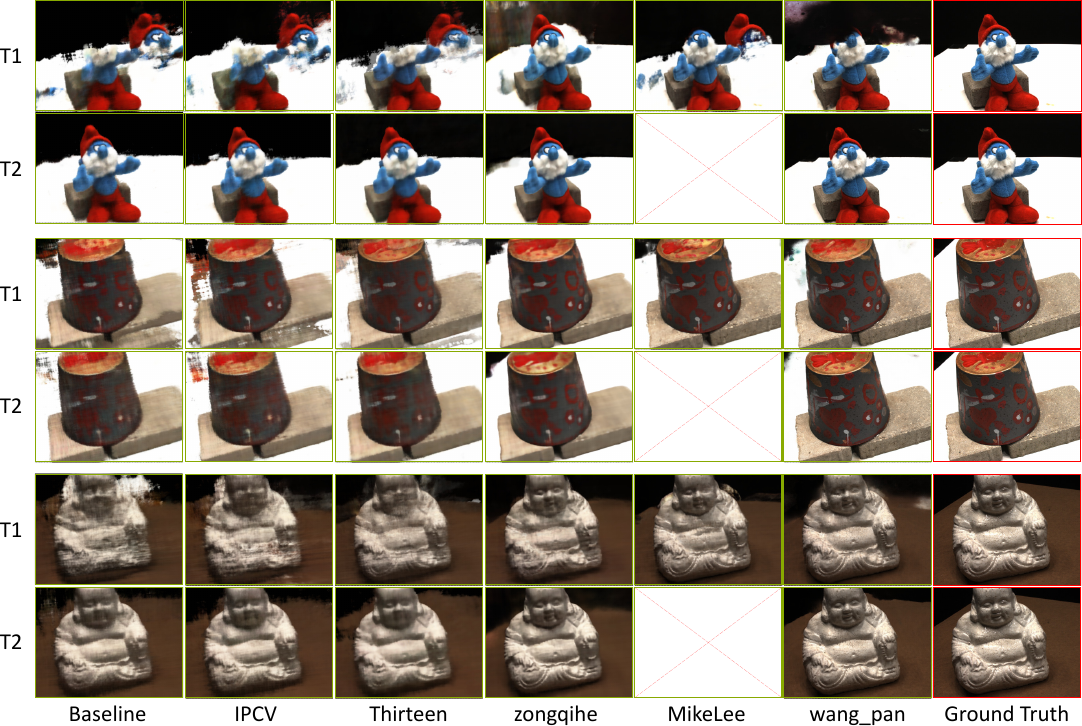}
    \caption{Test set results on the DTU dataset for Track 1 (T1) and Track 2 (T2).}
    \label{fig:results_dtu_test}
\end{figure}

In Figures \ref{fig:results_syn_test} and \ref{fig:results_dtu_test} we observe the visualisations of views generated by all the methods. Notably, in Figure~\ref{fig:results_syn_test} we can observe a higher quality of detail reconstruction for the winning solution for the SpaRe dataset. Observe the sharp writing on the side of the snow truck model and the detailed hinges on the chessboard, both more blurry for the other competitors. Similarly, for DTU (Figure~\ref{fig:results_dtu_test}) we observe a sharper reconstruction of the object by \textit{wang\_pan}. namely, Papa Smurf's plush texture, the graphics on the bucket, and the stone texture on the statue.

Notably, all teams improved upon the baseline solution which was an off-the-shelf implementation of FreeNeRF \cite{yang2023} trained at 4$\times$ downsampled resolution for computational efficiency and upsampled with bilinear interpolation for evaluation. The winning solution improved over the baseline by a large margin of $3.39dB$ in masked PSNR.

\subsubsection{Track 2}

In the final classification for Track 2 (Test phase - Table \ref{tab:track2test}) we observe a larger gap between the winner and the runner-up solutions. The winning team \textit{wang\_pan} achieved a high score of $24.51dB$ in masked PSNR leading over the second placed method by $2.92dB$. We observe a similar trend in perception-oriented metrics as well (SSIM-M of $0.784$ and LPIPS-M of $0.262$ with $0.135$ and $0.254$ advantage over the runner-up respectively).

In Figures \ref{fig:results_syn_test} and \ref{fig:results_dtu_test} we can see that even though the setting remains very challenging, Track 2 with 9 views poses fewer problems to the proposed algorithms than Track 1 with only 3 input views. With more input views, the ambiguity of the underlying 3D information is decreased which is reflected in the qualitative results. We observe typically better reconstruction around the edges of the object (see the snow truck in Figure\,\ref{fig:results_syn_test}, or the statue silhouette in Figure\,\ref{fig:results_dtu_test}). We also notice much fewer artefacts in the reconstruction, \eg continuity in Lego (Fig.\,\ref{fig:results_syn_test}) or Papa Smurf (Fig.\,\ref{fig:results_dtu_test}) geometries. We also notice differences between the classified solutions. In Figure\,\ref{fig:results_syn_test} we observe sharper reconstructions for \textit{wang\_pan} and \textit{Thirteen}, which is reflected in the respective scores for SpaRe dataset (PSNR-M: $24.46dB$ and $23.04dB$). We can see clear writing on the side of the snow truck or clear edges of the Lego bricks. Similarly, Figure\,\ref{fig:results_dtu_test} reflects the corresponding results on the DTU portion of the data, where \textit{wang\_pan} provides the sharpest images (note plush and stone textures, and the painting on the bucket), followed by \textit{zongqihe}. Notably, the ranking of the 2\textsuperscript{nd} (\textit{Thirteen}) and 3\textsuperscript{rd} (\textit{zongqihe}) places differed between SpaRe synthetic data and DTU data.

With the slightly easier task, the differences in the challenge participants' solutions on average were not as large with respect to the baseline as in Track 1. However, the winner achieved a margin of improvement of $4.08dB$ above the baseline FreeNeRF in masked PSNR.

\section{Conclusions}
This paper reviews the experimental set-up, methods, and results of the AIM Challenge on Sparse Neural Rendering held in conjunction with ECCV 2024. The problem set-up focuses on producing novel view synthesis of a scene given a sparse set of posed input images. The challenge is composed of two tracks: 3 input images in Track 1, and 9 input images in Track 2. Participants are asked to optimise PSNR with respect to the ground-truth images computed within an object mask. The dataset for the challenge is a combination of the SpaRe~\cite{Nazarczuk2024_dataset} (synthetic renderings from high-quality assets) and the DTU MVS~\cite{aanaes2016} (real captured images) datasets. Participants had access to a training set of 82 scenes, and submitted results on the validation set during the Development phase, and on the test set during the Final phase. A total of 5 teams submitted final results and factsheets in the Final phase. The submitted models obtained substantial improvements over existing baselines, with effective and varied solutions. The goal of this challenge is to standardise evaluation on sparse neural rendering models, and to stimulate future research in this field.

\subsection*{Acknowledgements}
This work was partially supported by the Humboldt Foundation. We thank the AIM 2024 sponsors: Meta Reality Labs, KuaiShou, Huawei, Sony Interactive Entertainment and University of W\"urzburg (Computer Vision Lab).

\appendix
\section{Teams and Affiliations} \label{sec:affiliations}

\subsection*{Sparse Neural Challenge Organisers}
\textbf{Members:} Michal Nazarczuk\textsuperscript{1} \email{[michal.nazarczuk1@huawei.com]}, Sibi Catley-Chandar\textsuperscript{1}, Thomas Tanay\textsuperscript{1}, Richard Shaw\textsuperscript{1}, Eduardo Pérez-Pellitero\textsuperscript{1}, Radu Timofte\textsuperscript{2}

\noindent
\textbf{Affiliations:} \textsuperscript{1}Huawei Noah’s Ark Laboratory, \textsuperscript{2}University of Würzburg

\subsection*{\textit{wang\_pan}}
\textbf{Members:} Xing Yan\textsuperscript{1}, Pan Wang\textsuperscript{1}, Yali Guo\textsuperscript{1}, Yongxin Wu\textsuperscript{1}, Youcheng Cai\textsuperscript{2}, Yanan Yang\textsuperscript{1}

\noindent
\textbf{Affiliations:} \textsuperscript{1}Hefei University of Technology, \textsuperscript{2}University of Science and Technology of China

\subsection*{\textit{MikeLee}}
\textbf{Members:} Junting Li\textsuperscript{1}, Yanghong Zhou\textsuperscript{1,2}, P. Y. Mok\textsuperscript{1,2}

\noindent
\textbf{Affiliations:} \textsuperscript{1}The Hong Kong Polytechnic University, \textsuperscript{2}Research Centre of Textiles for Future Fashion

\subsection*{\textit{zongqihe}}
\textbf{Members:} Zongqi He, Zhe Xiao, Kin-Chung Chan, Hana Lebeta Goshu, Cuixin Yang, Rongkang Dong, Jun Xiao, Kin-Man Lam

\noindent
\textbf{Affiliations:} Department of Electrical and Electronic Engineering, The Hong Kong Polytechnic University

\subsection*{\textit{Thirteen}}
\textbf{Members:} Jiayao Hao, Qiong Gao, Yanyan Zu, Junpei Zhang, Licheng Jiao, Xu Liu 

\noindent
\textbf{Affiliations:} Intelligent Perception and Image Understanding Lab, Xidian University

\subsection*{\textit{IPCV}}
\textbf{Members:} Kuldeep Purohit

\noindent
\textbf{Affiliations:} Google, Mountain View, USA

\bibliographystyle{splncs04}
\bibliography{main}
\end{document}